\documentclass[10pt,twocolumn,letterpaper]{article}

\usepackage{cvpr}
\usepackage{times}
\usepackage{epsfig}
\usepackage{graphicx}
\usepackage{amsmath}
\usepackage{amssymb}
\usepackage{caption}  
\usepackage{subcaption}
\def\etal{{\em et al.~}}

\usepackage[pagebackref=true,breaklinks=true,letterpaper=true,colorlinks,bookmarks=false]{hyperref}

\cvprfinalcopy 

\def\cvprPaperID{73} 

\ifcvprfinal\fi
\begin{document}

\title{EgoReID: Cross-view Self-Identification and Human Re-identification in Egocentric Videos}
\title{EgoReID: Cross-view Self-Identification and Human Re-identification in Egocentric and Surveillance Videos}

\author{Shervin Ardeshir\\
{\tt\small ardeshir@cs.ucf.edu},\\
\and
Sandesh Sharma\\
{\tt\small tsandesh23@gmail.com}\\
\and
Ali Borji\\
{\tt\small aborji@crcv.ucf.edu}\\
\and
Center for Research in Computer Vision (CRCV)\\
University of Central Florida, Orlando, FL\\
}

\maketitle

\begin{abstract}
Human identification remains to be one of the challenging tasks in computer vision community due to drastic changes in visual features across different viewpoints, lighting conditions, occlusion, etc. Most of the literature has been focused on exploring human re-identification across viewpoints that are not too drastically different in nature. Cameras usually capture oblique or side views of humans, leaving room for a lot of geometric and visual reasoning. Given the recent popularity of egocentric and top-view vision, re-identification across these two drastically different views can now be explored. Having an egocentric and a top view video, our goal is to identify the cameraman in the content of the top-view video, and also re-identify the people visible in the egocentric video, by matching them to the identities present in the top-view video. We propose a CRF-based method to address the two problems. Our experimental results demonstrates the efficiency of the proposed approach over a variety of video recorded from two views. 
\end{abstract}

\section{Introduction}
Human re-identification has been studied extensively in the past in the computer vision community. Due to changes in parameters such as lighting condition, view-point, and occlusion, human re-identification is known to be a very challenging problem. However, in almost all attempts to solve human re-identification, the nature of the data is almost the same between the two cameras, leaving room for a lot of geometric reasoning. Both cameras usually capture humans with oblique or side view. This allows reasoning about the geometry of the targets and also the spatial correspondences among the pedestrians' bounding boxes such as their expected location of head, torso and legs, leaving room for better appearance based reasoning. On the contrary, in this paper, we aim to perform human re-identification from two very drastic viewpoints. Our objective is to design a framework capable of addressing two tasks. Our first task is to identify the camera holder in the content of the top-view video. Our second task is to re-identify people visible in the egocentric camera in the surveillance top-view video, assuming the camera-holder's identity is given. The first task is challenging since we do not have any information regarding the visual appearance of the camera holder. Our only input for solving this problem, is the content of his egocentric video. In this work, we solely focus on the content of the video in terms of the visual appearance and geometric position of detected human bounding boxes.   

\begin{figure}[t]
\centering
\includegraphics[width=\linewidth]{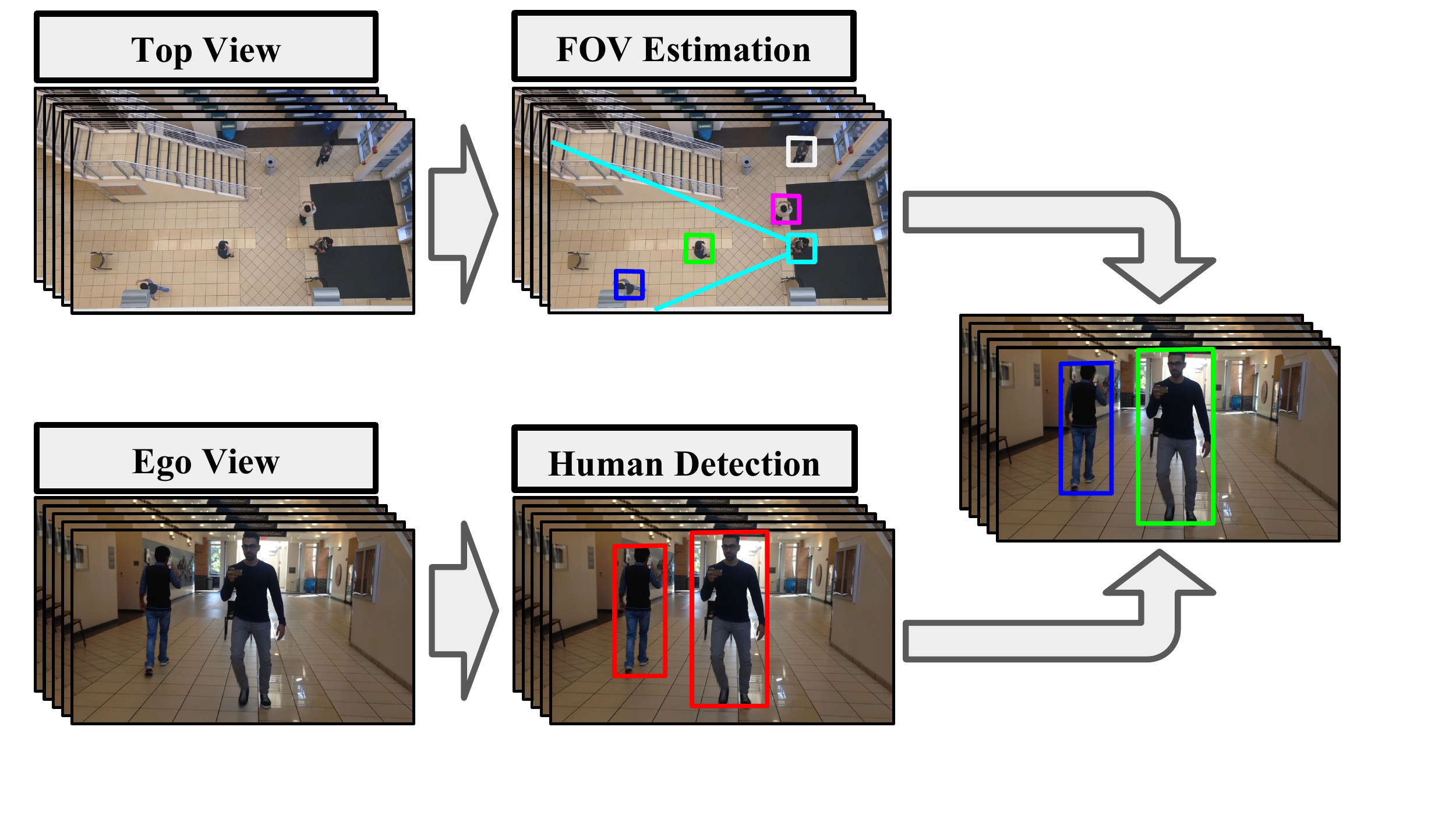}
\caption{Our framework consists of estimating field of view of the targets in the top-view video, and fusing that with the human detection results, in order to label the bounding boxes with the IDs present in the top-view video.}
\label{fig:main_framework}
\end{figure}

On one hand, egocentric vision has attracted a lot of interest during the past few years due to the abundance of affordable wearable cameras such as GoPro cameras and smart glasses. On the other hand, top- and oblique view videos have been very useful during the past decade due to the increasing affordability and capability of surveillance cameras, UAVs, and drones. These two sources of information provide very drastic view-points opening the door to a lot of exciting research aiming at relating these two sources of knowledge. 
 
Since top, side, and front views of a humans share very little appearance information, due to severe view-point changes and occlusion of the lower half of the human body, a direct comparison across these views will result in poor performance. Further, any sort of geometric reasoning requires some knowledge about the relative spatial position of the egocentric camera with respect to the top-view video. In other words, we need to know the identity and spatial location of the camera holder. This means that identification of the camera holder and the people in the content of the egocentric camera are related. Therefore, each of these two problems can be solved utilizing the solution of the other. This motivates a joint formulation that aims to jointly seek a labeling for the camera holder and people visible in the egocentric camera.

In this paper, we refer to identifying the camera-holder as \textit{self-identification}, and re-identifying the people visible in the egocentric video simply as \textit{re-identification}. We evaluate the cost of assigning each top-view identity to the camera holder, by fixing the self-identification label, and measuring the cost of the best possible re-identification labeling. Intuitively, if the self-identification is correct, the re-identification cost would be low, as the content of the camera is consistent with the expected content of the viewer in the top-view video. Assuming the self-identification identity is known, we estimate the initial re-identification labels based on some rough geometric reasoning between egocentric and top-view videos. Tracking the targets in the top-view video using a multiple object tracker, we can have an estimation of the camera-holder's field of view. We then infer an initial labeling for the humans visible in the egocentric video. We then modify the labeling by taking the visual appearance of the human detection results into account and enforcing visual consistency among different bounding boxes acquiring the same label. Once the re-identification labeling is finalized, the total cost of re-identification will be associated to the self-identification identity. Intuitively, the self-identification of an identity is evaluated based on the geometric consistency of the egocentric video with the identity's expected field of view in the top-view video. Our experiments show that these two cues provide complementary information and further improve our self-identification and re-identification results.   

To the best of our knowledge, the only previous work tackling the self-identification of egocentric viewers in top-view videos is \cite{ardeshir2016ego2top}. This approach requires a set of egocentric videos and heavily relies on pairwise relationships among egocentric cameras to reason about the assignment. Having only one egocentric video, the problem would be very difficult. In our experiments, we compare our results to \cite{ardeshir2016ego2top} as a baseline. In addition, \cite{ardeshir2016ego2top} does not address our second task, which is re-identifying people visible in the egocentric video. 

The rest of this paper is organized as follows. We selectively review the related work in Section 2. Our framework is described in Section 3. Experiments and results are explained in Section 4. Finally, Section 5 concludes the paper.

\section{Related Work}
In this section, we review related works in the areas of human re-identification and egocentric vision.
\subsection{Person Re-identification}
Person re-identification has been studied heavily during the past few years in the computer vision community \cite{reidCPS,reidReimannian,reidSDALF}. The objective here is to find and identify people across multiple cameras. In other words, who is each person present in one static camera, in another overlapping or non-overlapping static camera? The main cue in human re-identification is visual appearance of humans, which is absent in egocentric videos. 

Tasks such as human identification and localization in egocentric cameras have been studied in the past. \cite{egoHeadMotion} uses the head motion of an egocentric viewer as a biometric signature to determine which videos have been captured by the same person. In \cite{egoSurfing}, authors identify egocentric observers in other egocentric videos using their head motion. 
The work of \cite{egoFOVLocalization} localizes the field of view of an egocentric camera by matching it against a reference dataset of videos or images, such as Google street view images. 
Landmarks and map symbols have been used in~\cite{egoWhereAmI} to perform self localization on a map. The study reported in \cite{chakraborty2016person} addresses the problem of person re-identification in a surveillance network of wearable devices, and \cite{zhengidentifying} performs re-identification on time-synchronized wearable cameras. The relationship between egocentric and top-view information has been explored in tasks such as human identification \cite{ardeshir2016ego2top,ardeshiregocentric}, semantic segmentation\cite{ardeshir2015geo}  and temporal correspondence\cite{ardeshirEgocentricMeets}. \cite{ardeshir2016egotransfer} also seeks an automated method for learning a transformation between motion features across egocentric and non-egocentric domains.

One of the popular approaches in recent years is using deep learning for person re-identification \cite{DBLP:journals/corr/YiLL14, 6909421, Ahmed_2015_CVPR, Cheng_2016_CVPR, DBLP:journals/corr/VariorHW16, DBLP:journals/corr/VariorSLXW16}. Yi~\etal \cite{DBLP:journals/corr/YiLL14} uses "siamese" deep neural network for performing re-identification. The method proposed by Ahmed et al. \cite{Ahmed_2015_CVPR} uses improved deep neural network architectures to determine whether input pairs of images match. Cheng. et al. \cite{Cheng_2016_CVPR} uses a multi-channel CNN model in a metric learning based approach. Chen. et al. \cite{Chen_2016_CVPR} used spatial constraints for similarity learning for person re-identification, and combines local and global similarities. Cho. et al. \cite{Cho_2016_CVPR} uses a multi-pose model to perform re-identification. Matsukawa. et al. \cite{Matsukawa_2016_CVPR} uses a region descriptor based on hierarchical Gaussian distribution of pixel features for person re-identification.

\subsection{Egocentric Vision}
Visual analysis of egocentric videos has recently became a hot research topic in computer vision~\cite{egoKanade,egoEvolutionSurvey}, from recognizing daily activities \cite{egoDailyAction,egoActionFathi} to object detection \cite{egoObjectDetection}, video summarization \cite{egoVideoSummarization}, and predicting gaze behavior~\cite{egoli2013learning,egoPolatsekNovelty,Borji2014look}.
Some studies have addressed relationships between moving and static cameras. 
Interesting works reported in \cite{egoMobileFixedObjectDetection,egoMobileFixedMasterSlave} have explored the relationship between mobile and static cameras for the purpose of improving object detection accuracy. \cite{egoExo} fuses information from egocentric and exocentric vision (third-person static cameras in the environment) with laser range data to improve depth perception and 3D reconstruction. Park {\em et al.} \cite{egoPredictingGaze} predict gaze behavior in social scenes using first-person and third-person cameras. Soran {\em et al.},~\cite{soran2014action} have addressed action recognition in presence of an egocentric video and multiple static videos.

\section{Framework}
\begin{figure}
\centering
\includegraphics[width=1\linewidth]{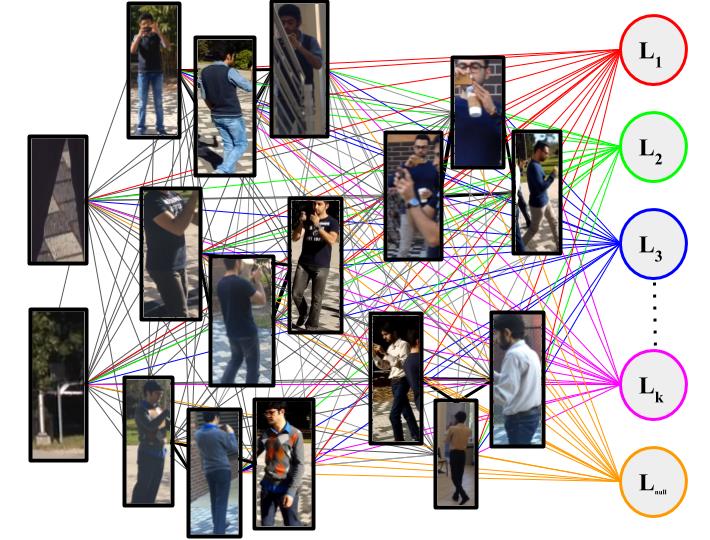}
\caption{Graph construction: Each human detection bounding box is a node. Each of the labels correspond to one of the identities in the top-view video. Edges in the graph encode distance in terms of visual features and satisfying our mentioned spatio-temporal constraints.}
\label{fig:graph}
\end{figure}

The block diagram of our proposed method can be seen in figure \ref{fig:main_framework}. Given an egocentric video and a top-view video containing $n_{top}$ identities, we run human detection on the egocentric video \cite{sighthound} which will provide us a set of bounding boxes $D=\{d_1,d_2,...d_{n_d}\}$. Re-identification is defined as labeling the human detection bounding boxes by assigning them to the viewers in the top-view video labeled with $L=\{l_1,l_2,...l_{n_{top}}\}$ where $n_{top}$ is the number of people visible in the top-view video. We evaluate the re-identification labeling cost, assuming each of the viewers in the top-view video to be the camera-holder. We then rank the viewers in the top-view video based on their likelihood of being the cameraman. We evaluate the performance of that ranking and compute the effects of different parts of our formulation. We also evaluate the human re-identification labeling accuracy, assuming the correct egocentric ID is given. Assuming the self identification label for the egocentric video is $l_s \in \{1,2,...,n_{top}\}$, we seek the best set of re-identification labels $L_r \in \{ 1,2,3,...,n_{top} \}^{n_{d}}$. 
\begin{equation}
L_r,l_s = \underset{\mathbf{L_r,l_s}}{\operatorname{argmin}} \ C(L_r|l_s).
\end{equation}
Our framework for computing $C(L_r|l_s)$ contains two main steps.
First, using $l_s$, we compute a set of initial labeling solely based on geometric configuration of targets in the top-view video. We then penalize visually similar bounding boxes to acquire different labels. We model our objective using a graph $G(V,E)$, where each node $v_i$ is a human detection bounding box as shown in figure \ref{fig:graph}. Each node is eventually going to receive a label by being matched to one of the top-view identities($l_1,l_2,...,l_{n_{top}}$). Each edge also, captures the cost of assigning the same label to the nodes on its two ends. The details of each of the two steps are explained in the following sections.

\subsection{Geometric Reasoning from Top-view}
\begin{figure}
\centering
\includegraphics[width=1\linewidth]{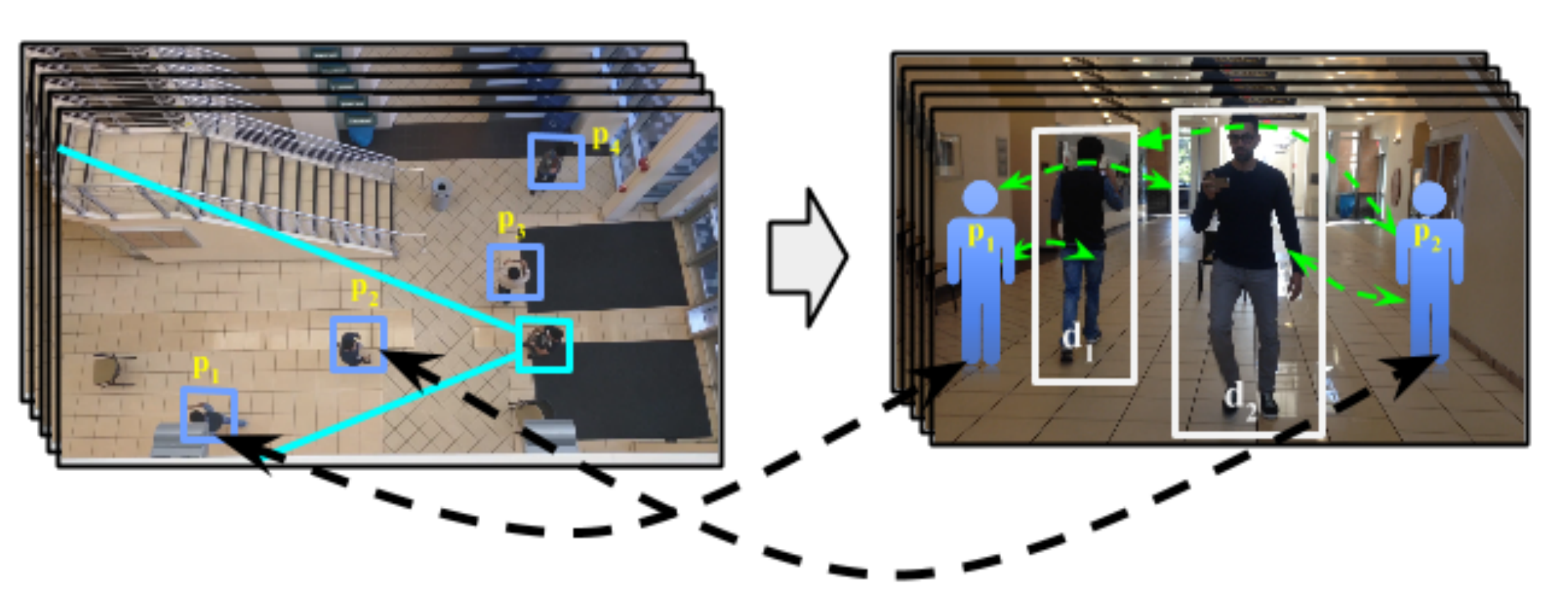}
\caption{Projecting the information from the top-view video to the egocentric space. As shown in the figure (left), two identities are visible in the field of view of the egocentric viewer. Using their orientation with respect to the viewer's direction of movement in the top-view, a rough estimate of their location in the content of his egocentric video can be computed.}
\label{fig:geometry}
\end{figure}

As mentioned before, we evaluate all identities present in the top-view video in terms of being the egocentric camera-holder independently. We then compare their recommended labeling costs to perform self-identification. Having $n_{top}$ identities present in the top-view, we compute $n_{top}$ labeling costs $C(L_r|l_s=1), C(L_r|l_s=2), ..., C(L_r|l_s=n_{top})$, which captures how visually and geometrically consistent their recommended labeling is for the re-identification task. To perform geometric reasoning from top-view, similar to \cite{ardeshir2016ego2top} we perform multiple object tracking on the provided top-view bounding boxes. Knowing the direction of motion of each trajectory at each moment, we employ the same assumptions as in \cite{ardeshir2016ego2top} and estimate the head direction of each of the top-view viewers by assuming that the viewers tend to look straight ahead in majority of times. Also, not having access to the intrinsic parameters of the egocentric video such as focal length and sensor size, we assume a fixed angle and therefore estimate the field of view of each viewer as illustrated in figure \ref{fig:geometry}. As a result, we can determine which identities are expected to be visible in the field of view of each viewer. Thus, given the self-identity ($l_s$), we can acquire a set of suggested re-identification labeling from the top-view video. 

As shown in figure \ref{fig:geometry}, using the relative location and orientation of the visible top-view bounding boxes, we can estimate its spatial location in $x$ axis in the egocentric video content as $\frac{d_1w}{2d_2}$ relative to the center of the frame. In the previous term, $d_1$ is the spatial distance between the top-view bounding box to the orientation ray, and $d_2$ is the spatial distance between the orientation ray and the border of the field of view cone as depicted in figure \ref{fig:geometry}. Also $w$ is the width of the egocentric video and therefore $\frac{d_1w}{sd_2}$ will encode the $x$ axis distance of the projection of the top-view bounding box in the content of the egocentric video relative to the center of the frame. Estimating a projection for each individual top-view visible bounding box, we will have a set of image $x$-axis coordinates $p_1,p_2,...,p_{k}$, where $k$ is the number of visible top-view bounding boxes(2 in the example shown).\\

To capture the cost of assigning each detection to a projection, we compute the distance between each projection $p_i$ with each human detection center $d_j$, and form a $n_p \times n_d$ matrix, containing the matching probability between a projection-detection pair. The distance matrix is computed as $M(i,j)=dist(p_i,d_j)$. In order to maintain a notion of probability, we enforce that each projection should match to one and only one detection, and at the same time, each detection should match to one and only one projection. We also perform bi-stochastic normalization on matrix $M$ to ensure $M\mathbf{1}\approx\mathbf{1}$ and $\mathbf{1}M\approx\mathbf{1}$. Bi-stochastic normalization could be done simply by an iterative row-wise and column-wise normalization up to reaching within a convergence error.   

We define the unary cost of a node by evaluating the cost of assigning each of the top-view labels to that node. Therefore, having $n_{top}$ people visible in the top-view video, we can represent the unary cost of a node (human detection bounding box) $i$ ($C_{i}^u$) to be a $1 \times n_{top}$ vector, where: 
\begin{equation}
C^u(L(i)=l_j)=
\begin{cases}
M(i,l_j)\textit{\ j is visible in i's FOV}\\
1 \textit{\ otherwise}.
\end{cases}
\end{equation}
\subsection{Visual Reasoning}
Having $n_{d}$ human detection bounding boxes and therefore $n_d^2$ edges containing cost of associating the same label to the two nodes, the cost $C_{v}$ contains the euclidean distances computed on visual features that we extract from the bounding boxes. For computing this cost, we extract visual features including color histogram, LBP texture \cite{ojala1996comparative}, and CNN features using the VGG-19 deep network\cite{simonyan2014very}, from the human detection bounding boxes. We then use $L2$ normalization, and PCA on the CNN features and reduce their dimensionality to 100. Features are concatenated and used to represent the visual information in a human detection bounding box. 
\subsection{Spatio-temporal Reasoning}
We also incorporate a spatiotemporal cost in the graph edges $\mathbf{C}_{st}$ capturing some spatial and temporal constrains on bounding boxes. These constraints are defined as the following:\\

\noindent\textbf{Constraint 1:} Each pair of bounding boxes present in the same frame cannot belong to the same person. Therefore the binary cost between any pair of co-occurring bounding boxes is set to infinity.\\

\begin{figure*}[h]
\centering
\includegraphics[width=\linewidth]{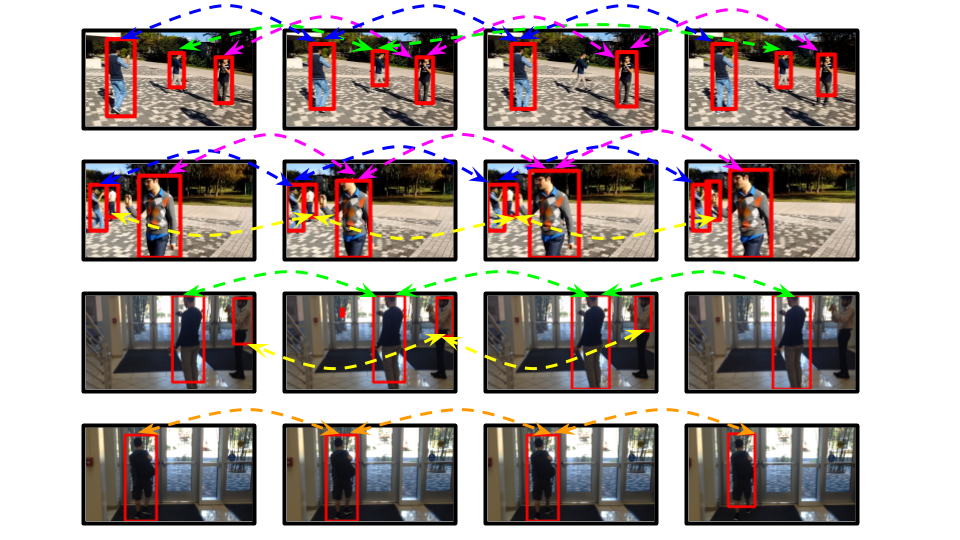}
\caption{A few examples of the spatio-temporal overlaps which we incorporate in our pairwise distances. Intuitively, we generate some very short term, but very confident tracklets whenever possible, in order to enhance the consistency in our labeling.}
\label{fig:IOU}
\end{figure*}

\noindent\textbf{Constraint 2:} If two bounding boxes have a very high overlap in temporally nearby frames, their binary cost will be alleviated, as they would probably belong to the same identity. An example is shown in figure \ref{fig:IOU}.

\begin{figure*}
\begin{center}
   \includegraphics[width=1\textwidth]{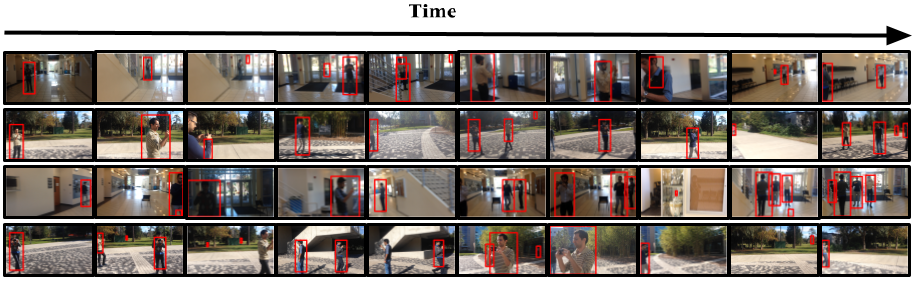}
   \caption{A few examples of the video sequences alongside with the human detection results. Each row represents sample frames from one sequence.}	
\end{center}
\end{figure*}

\subsection{Handling Temporal Misalignment}
As mentioned in \cite{ardeshiregocentric}, a perfect time-alignment between the egocentric and top-view videos is not available and therefore our framework should be able to handle temporal misalignment between the sources. In order to cope with that, we compute the labeling cost for different relative time delays between the egocentric and top-view video and assign the lowest cost set of labels to the human-detection bounding boxes.  
Time delays will alter the initial labeling and the unary costs as frame $f$ in the egocentric video will correspond to frame $f+\tau$ in the top-view video. Therefore, the labels from the visible humans at frame $f + \tau$ will be propagated to the human detection bounding boxes in frame $f$ in the egocentric video. As a result, labeling cost would be a matter of the time-delay that we assign.

\begin{equation}
L = \underset{\mathbf{\tau, L}}{\operatorname{argmin}} \ C_{tot}(\tau,L).
\label{fig:joint_opt}
\end{equation}

\subsection{Fusion}
In order to combine our initial labeling with the visual content of the human detection bounding boxes, we use graph cuts \cite{boykov2001fast} to select the minimum cost labeling. The unary geometry based term ($C_{g}$) comes from the top-view suggested initial labeling, and the second and third term includes the binary costs for assigning different bounding boxes to different labels.

\begin{equation}
C_{tot}(L,\tau) = \sum_{i=1}^{k} C_{g}(L_i,\tau) + \sum_{j=1,j\neq i}^{k} C_{v}(L_i,L_j) + C_{st}(L_i,L_j)
\end{equation}
Intuitively, we initialize a labeling with different time-delays using the geometric reasoning, and then enforce visual and spatiotemporal consistency among the similarly labeled nodes by incorporating the binary costs. Our experiments show that the fusion will further improve the re-identification accuracy. At the end, we pick the configuration with the lowest labeling cost as in equation \ref{fig:joint_opt}.

\begin{figure*}[t]
\begin{center}
   \includegraphics[width=1\textwidth]{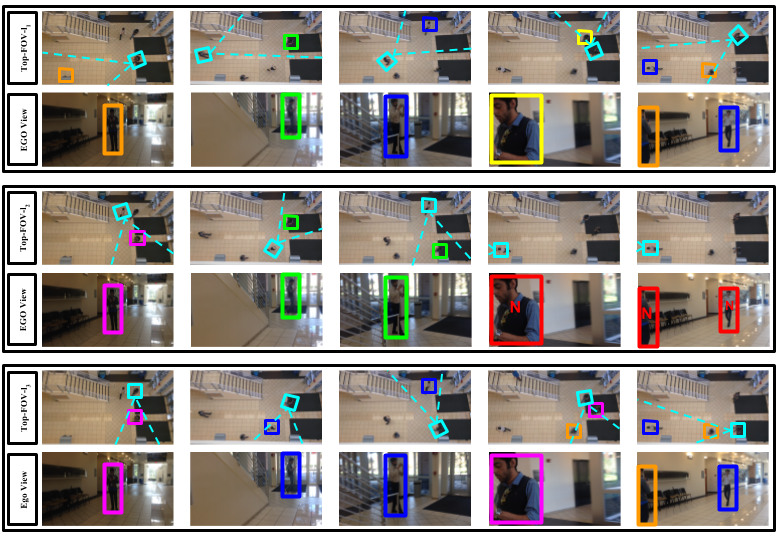}
   \caption{The labeling resulting from each of the self identification IDs. Dependeing on the self identification, the top-view field of view and therefore the content of it would be different, resulting in a different labeling. First example is the re-identification results, obtained, if the self identification label is correct.}	
\end{center}
\end{figure*}

\section{Experimental Results}
In this section, we will explain our experimental setup and dataset that will used for evaluating the two tasks of self-identification, and human re-identification. We then evaluate the performance of our proposed method over two tasks and analyze the performance.

\subsection{Dataset}
We use the first 10 sequences of the dataset used in \cite{ardeshir2016ego2top}. The dataset contains test cases of videos shot in different indoor and outdoor environments. Each test case contains one top-view video and several egocentric videos captured by the people visible in the top-view camera. We annotated the labels for the human detection bounding boxed for each video and evaluated the accuracy for re-identification and self-identification. The first 10 sets contain 37 egocentric videos and 10 top-view videos. Number of people visible in the top-view cameras varies from 3 to 10, and lengths of the videos vary from 1019 frames (33.9 seconds) up to 3132 frames (110 seconds). 

\subsection{Evaluation}
We evaluate our framework in terms of egocentric self-identification within a top-view video, and also in terms of cross-view human re-identification. 

\subsubsection{Self-identification}
\begin{figure}[h]
\centering
\includegraphics[width=0.8\linewidth]{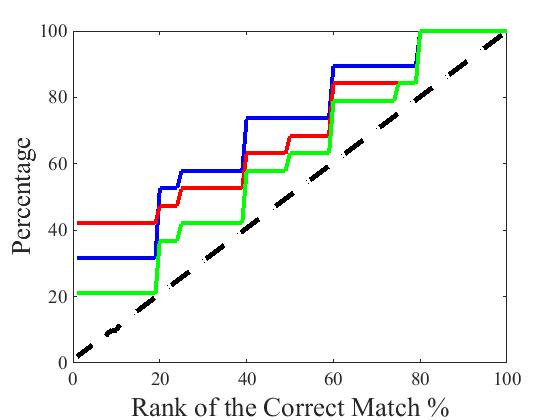}
\caption{Cumulative matching curve illustrating the performance of our proposed method, before fusion and only using the geometric reasoning (blue), and after fusion using the visual and spatiotemporal reasoning (red). The green curve shows the performance of the baseline method proposed in \cite{ardeshir2016ego2top} using cross correlation of time series constructed from the number of humans present in the video. Our proposed method clearly outperforms the baselines.}
\label{fig:cmc}
\end{figure}

For each egocentric video, the viewers visible in the top-view video are ranked and self-identification performance is evaluated by computing the area the cumulative matching curve (CMC) as illustrated in figure \ref{fig:cmc}. We also compared the self-identification accuracy with that of \cite{ardeshir2016ego2top} where they only use results of human detection for performing identification of the camera holder. The reason behind the cumulative matching curves having jumps in values and non-smooth transitions, relies on the fact that there are only a few people visible in the top-view video and therefore the normalized rank of the correct match could only obtain limited number of values (e.g. $\{0.2,0.4,...,1\}$ for ranks 1 to 5 when 5 people are visible in the top-view video).  

\subsubsection{Cross-view Human Re-identification}
Assuming the egocentric identity is known, the labeling accuracy is computed for the bounding boxes visible in the content of the egocentric video. The labeling accuracy is evaluated for the initial labeling suggested by the top-view video, and also after fusing that with the visual similarities of the bounding boxes.
\begin{figure}[h]
\begin{center}
\includegraphics[width=0.5\textwidth]{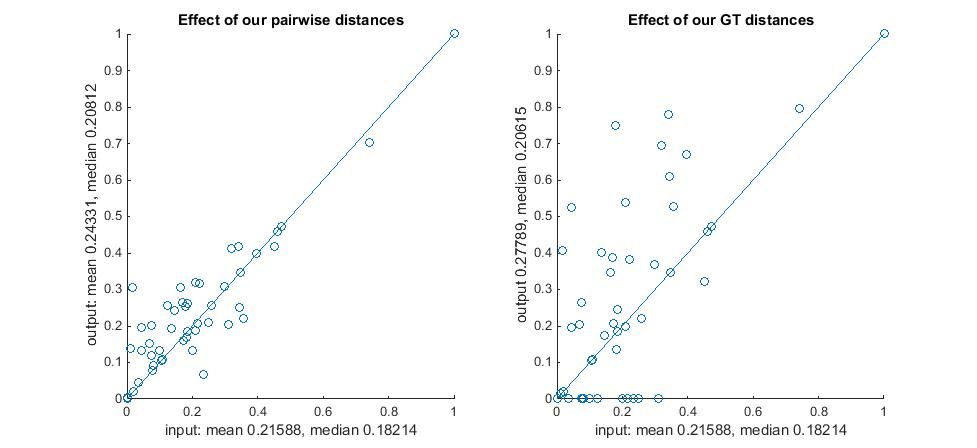}
\caption{Left: Scatter plot capturing the effect of incorporating visual and spatiotemporal consistency. It shows that simply using the geometric reasoning from the top view video leads to inferior re-identification accuracy compared to our proposed fusion framework. Right: shows the same comparison with the difference that the binary similarities were formed using the groundtruth labels of the re-identificatin bounding box. This will give us an upper bound for what is possible to achieve using our initial labeling and an ideal pairwise cost.}
\label{fig:scatter}
\end{center}
\end{figure}
   
\begin{figure}
\begin{center}
   \includegraphics[width=0.4\textwidth]{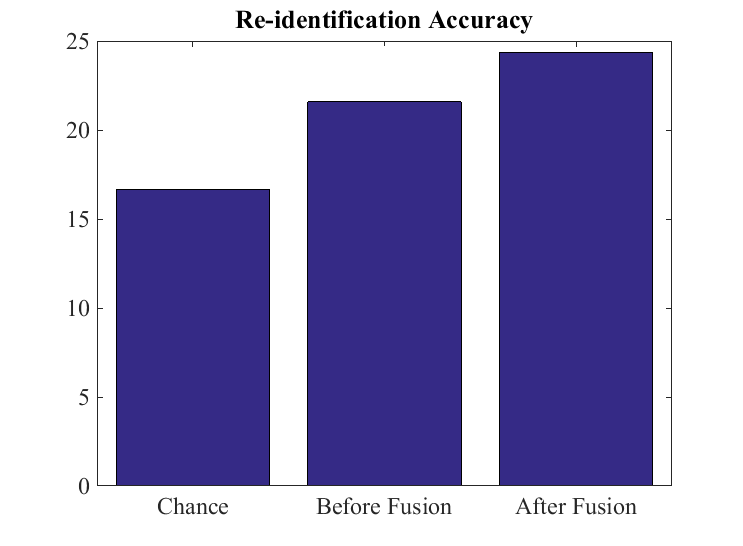}
   \caption{The re-identification labeling accuracy. It can be observed that simply using the geometrical reasoning from top-view we can achieve re-identification labeling accuracy which outperforms chance. Also, incorporating the pairwise similarities, will further improve the re-identification labeling.}	
\end{center}
\end{figure}

\section{Conclusion}
In this work we studied the problem of human re-identification and self-identification in egocentric videos, by matching them to a reference top-view surveillance video. Our experiments show that both self-identification and re-identification is possible in a unified framework. If self-identification is given, re-identification can be done using the some rough geometric reasoning from top-view and enforcing visual consistency.  

For future, a more general case of this
problem can be explored such as assigning multiple egocentric
viewers to viewers in multiple top-view cameras. Also,
other approaches can be explored for solving the introduced
problem or slight variations of it (e.g., supervised methods
for understanding the unary and pairwise relationships).
Further, other computer vision techniques such as visual
odometry can be explored for relating the two sources. We
attempted to approach this problem using odometry at first,
however, the results were not accurate perhaps due to a
lot of jitter in egocentric videos. Nonetheless, this can be
another potential direction for further research in the future.

{\small
\bibliographystyle{ieee}
\bibliography{Thesis_bib}
}

\end{document}